\documentclass[runningheads]{llncs}
\usepackage[T1]{fontenc}
\usepackage{graphicx}
\usepackage{wrapfig}
\usepackage{cite}
\usepackage{algorithm} 
\usepackage{algpseudocode} 
\usepackage{subfigure}
\usepackage{subcaption}
\usepackage[inkscapelatex=false]{svg}
\usepackage{amsmath}
\usepackage{amssymb}
\usepackage{comment}
\usepackage{booktabs} 

\begin{document}
\title{ABBA-VSM: Time Series Classification using Symbolic Representation on the Edge}
\titlerunning{ABBA-VSM}
%
 \author{Meerzhan Kanatbekova \orcidID{0009-0008-4847-1107} 
 \and
 Shashikant Ilager \orcidID{0000-0003-1178-6582} 
 \and
 Ivona Brandic \orcidID{0000-0001-7424-0208}
 }

 \authorrunning{M. Kanatbekova et al.}
%
\institute{TU Wien, Vienna, Austria \\
\email{\{meerzhan.kanatbekova, shashikant.ilager, ivona.brandic\}@tuwien.ac.at}}
%
\maketitle              
\begin{abstract}

In recent years, Edge AI has become more prevalent with applications across various industries, from environmental monitoring to smart city management. Edge AI  facilitates the processing of   Internet of Things (IoT) data and provides privacy-enabled and latency-sensitive services to application users using Machine Learning (ML) algorithms, e.g., Time Series Classification (TSC). However, existing   TSC algorithms require access to full raw data and demand substantial computing resources to train and use them effectively in runtime. This makes them impractical for deployment in resource-constrained Edge environments. To address this, in this paper, we propose an Adaptive Brownian Bridge-based Symbolic Aggregation Vector Space Model (ABBA-VSM). It is a new TSC model designed for classification services on  Edge. Here, we first \textit{adaptively} compress the raw time series into symbolic representations, thus capturing the changing trends of data. Subsequently, we train the classification model directly on these symbols.
ABBA-VSM reduces communication data between IoT and Edge devices, as well as computation cycles, in the development of resource-efficient TSC services on Edge.
We evaluate our solution with extensive experiments using datasets from the UCR time series classification archive. The results demonstrate that the ABBA-VSM achieves up to 80\% compression ratio and 90-100\% accuracy for binary classification. Whereas, for non-binary classification, it achieves an average compression ratio of 60\% and accuracy ranging from 60-80\%.

\keywords{Edge Computing \and EdgeAI \and Time Series Classification \and Data Compression \and Symbolic Representation}
\end{abstract}
\section{Introduction}
The number of Internet of Things (IoT) devices worldwide is anticipated to experience a significant increase, nearly doubling from 15.9 billion in 2023 to over 32.1 billion by 2030, according to data from Statista (2023). They produce a massive amount of data, and it is expected to reach around 85 zettabytes (ZB) by 2025~\cite{statistaConnectionsWorldwide, aslanpour2021wattedge, alam2023reliable}. All these are driven by the widespread utilization of IoT technology and services across various industries, from environmental monitoring~\cite{tung2020survey} to smart city management~\cite{liu2024new}. Many of the IoT applications leverage Machine Learning (ML) algorithms in their pipeline to provide data-driven smart services. Traditionally, IoT data has been transmitted to, stored, and processed in the cloud~\cite{shukla2023improving}. However, the growing demand for latency-sensitive application services and privacy requirements has introduced a new paradigm called Edge AI. Edge AI provides limited computing resources to design and deploy applications at the network  Edge~\cite{ilager2023data, sabovic2023towards, rosero2023hybrid}. 

Among many ML algorithms, Time Series Classification (TSC) is a widely applicable popular method that predicts a class label of a given Time Series (TS). TSC has various real-world applications in many domains, such as smart city management tasks~\cite{liu2024new}, and environmental monitoring~\cite{tung2020survey}, among others. 

The state-of-the-art TSC algorithms primarily focus on accuracy, demonstrating the capability to achieve high levels of accuracy across different datasets. However, these algorithms face a challenge in classifying large datasets in resource-constrained environments such as \textit{Edge}. First, off-the-shelf TSC algorithms require sending full raw data from IoT to Edge devices. This is infeasible in the Edge environment since communication between IoT and Edge devices is expensive,  some studies have shown that communication costs up to 80\% of energy in IoT devices~\cite{aslanpour2021wattedge}. Moreover, Edge devices often have limited computing and memory resources to process the raw data and train new TSC models~\cite{azar2019energy}.   Furthermore, data from many applications has to be processed in near real-time~\cite{ilager2023data}. Thus, it is necessary to develop a resource-efficient TSC method to handle the growing volume of IoT data generated and to manage Edge applications efficiently.

In response to these challenges, various strategies have been proposed to optimize the data processing at Edge~\cite{ilager2023data, 10.1007/978-3-031-39698-4_28, azar2019energy}. For instance, in~\cite{azar2019energy}, 
authors propose data compression at the IoT level and train the ML model on reconstructed data at the Edge. While such a method solves the problem of high data traffic between IoT and Edge, Edge's memory and computation constraints remain challenging. Consequently, Symbolic Representation (SR)~\cite{malinowski20131d, middlehurst2024bake} methods offer an alternative approach to reduce the data size of numerical time series data and perform analytics on reduced data.  The SR is a lossy data compression technique that partitions raw data into segments (chunks) and encodes them by symbols, creating a string of symbols. If required, symbols could be reconstructed back to the original time series with a controllable error rate.  Contrary to classical compression methods, SR preserves data semantics, allowing us to do data analytics directly on symbols. 

In this work, we explore how we can leverage SR to develop a time series classification model that is directly trained on symbols.  While some works~\cite{senin2013sax} have explored symbolic methods for TSC using  SAX-VSM, such approaches do not apply to adaptive (streaming IoT data) compression and latency-constrained applications. Our approach is feasible in an environment where IoT devices can compress the data and transfer the reduced data to the Edge, thus reducing communication and storage costs between IoT and Edge, and processing costs at the Edge. 

Therefore, we propose ABBA-VSM (\underline{A}daptive \underline{B}rownian \underline{B}ridge-based symbolic
\underline{A}ggregation
\underline{V}ector \underline{S}pace \underline{M}odel), an adaptive approach for TSC using SR. It consists of two main components: \texttt{compression} and \texttt{classification}. We first present the adaptive time series compression inspired by the Brownian bridge to reduce the size of the raw TS~\cite{elsworth2020abba}, followed by encoding the compressed data as a string of symbols. Second, the Vector Space Model (VSM) is constructed to build a TSC model. Finally, the ABBA-VSM outputs the TSC model trained to classify the next set of data points. Our approach is adaptive to the non-stationary data, dynamically adapts to create accurate symbols, and provides an algorithm for TSC, which is directly trained on symbols.

In summary, the \textbf{key contributions} of the paper are:
\begin{itemize}
    \item We propose a new adaptive symbolic time series classification model for latency-constrained Edge applications, exploring its impact on memory and computation constraints.
    \item We empirically evaluate the proposed method on real-world datasets and compare it to non-adaptive baseline approaches.
    \item Our extensive experiments demonstrate that ABBA-VSM achieves 90-100\%  accuracy for binary classification datasets and achieves reasonable accuracy for multi-class classification.
\end{itemize}
The rest of the paper is organized as follows. Section \ref{sec:related_work} provides an overview of the existing symbolic TSC methods. Section \ref{sec:problem_formulation} describes the real-world application scenario. Section \ref{sec:abba_vsm} explains the proposed ABBA-VSM method in detail. Section \ref{sec:experiments} describes the experimental design and datasets used. Section~\ref{sec:results} discusses the empirical results. Finally, Section~\ref{sec:conclusion} presents the concluding remarks and potential future work. 

\section{Related Work} \label{sec:related_work}
Multiple techniques exist for TS reduction and classification, but only a few are designed to represent TS data symbolically. Here, we will provide an overview of existing symbolic data compression methods and symbolic time series classification algorithms, which are summarized in Table~\ref{tab:related_work}. 

\begin{wraptable}{r}{7.5cm}
    \vspace{-1\baselineskip}
    \caption{Comparison of most relevant works that use symbolic representation  for TS classification}
    \label{tab:related_work}
    \scriptsize
    \begin{tabular}{c|c|c|c}
    \toprule
    \textbf{Algorithm} &
    \begin{tabular}[c]{@{}l@{}}\textbf{Adaptive} \\ \textbf{reduction}\end{tabular} & 
    \begin{tabular}[c]{@{}l@{}}\textbf{Symbol} \\ \textbf{generation} \end{tabular} & 
    \begin{tabular}[c]{@{}l@{}}\textbf{Classification} \\ \textbf{on the Edge} \end{tabular} \\
    \midrule
    SAX-VSM ~\cite{senin2013sax}  & $\times$ & \checkmark & $\times$ \\
    BOSS ~\cite{schafer2015boss} & $\times$ & \checkmark & $\times$ \\
    WEASEL ~\cite{schafer2017fast} & $\times$ & \checkmark & $\times$  \\
    MrSQM  ~\cite{nguyen2021mrsqm} & $\times$ & \checkmark & $\times$ \\
    \textbf{ABBA-VSM} & $\checkmark$ & \checkmark & $\checkmark$ \\
    \bottomrule
    \end{tabular}
    \vspace{-1\baselineskip}
\end{wraptable}

The TSC techniques can be broadly categorized into two groups: \textit{full} time series-based methods and \textit{feature-based} methods~\cite{lin2012rotation}. \textit{Full} time series-based methods use a pointwise comparison of TS, for instance, with 1-NN Euclidean Distance (ED) or 1-NN Dynamic Time Warping (DTW). While these techniques are well suited for short TS, they are inefficient for long and noisy TS. Whereas \textit{feature-based} techniques compare features or segments generated from full TS. The common approach within this \textit{feature-based} group is the Bag-Of-Patterns (BOP) model~\cite{lin2012rotation}. Such models are built by breaking up a TS into segments representing discrete features, creating a bag of words from these features, and finally, building a histogram of feature counts as a basis for classification. 

Most BOP models employ Symbolic Aggregate Approximation (SAX) and Symbolic Fourier Approximation (SFA), which are commonly used methods for creating linear segments from TS~\cite{lin2012rotation, pham2010hot}. SAX partitions TS into segments of fixed length and then represents each segment by the mean of its values (i.e., a piece-wise constant approximation).  In contrast, SFA converts TS into symbolic representations using Fourier coefficients~\cite{schafer2012sfa}.

The Bag of Symbolic Fourier Approximation Symbols (BOSS) algorithm is based on the SFA method, which involves approximating the original data using the Discrete Fourier Transform and then discretizing the resulting coefficients using a technique called Multiple Coefficient Binning~\cite{schafer2015boss}. Multiple Representations Sequence Miner (MrSQM) offers four different feature selection strategies, including random feature selection, pruning the all-subsequence feature space, and random sampling of features~\cite{nguyen2021mrsqm}. Word ExtrAction for time SEries cLassification (WEASEL) is a TSC method that creates a large SFA words feature space, filters it with Chi-square feature selection, then trains a logistic regression classifier~\cite{schafer2017fast}. Symbolic Aggregate Approximation Vector Space Model (SAX-VSM) is a TSC method that creates a feature space of SAX words and then builds a classifier by building the weight matrix of SAX words~\cite{senin2013sax}. 

While these methods showcase the diversity in feature extraction and symbolic classification techniques for TS, they often fail to capture the changing trends in the data during real-time TS compression. 

\section{System Model} \label{sec:problem_formulation} 
\begin{wrapfigure}{l}{0.60\textwidth}
    \vspace{-2\baselineskip}
    \includegraphics[width=0.59\textwidth]{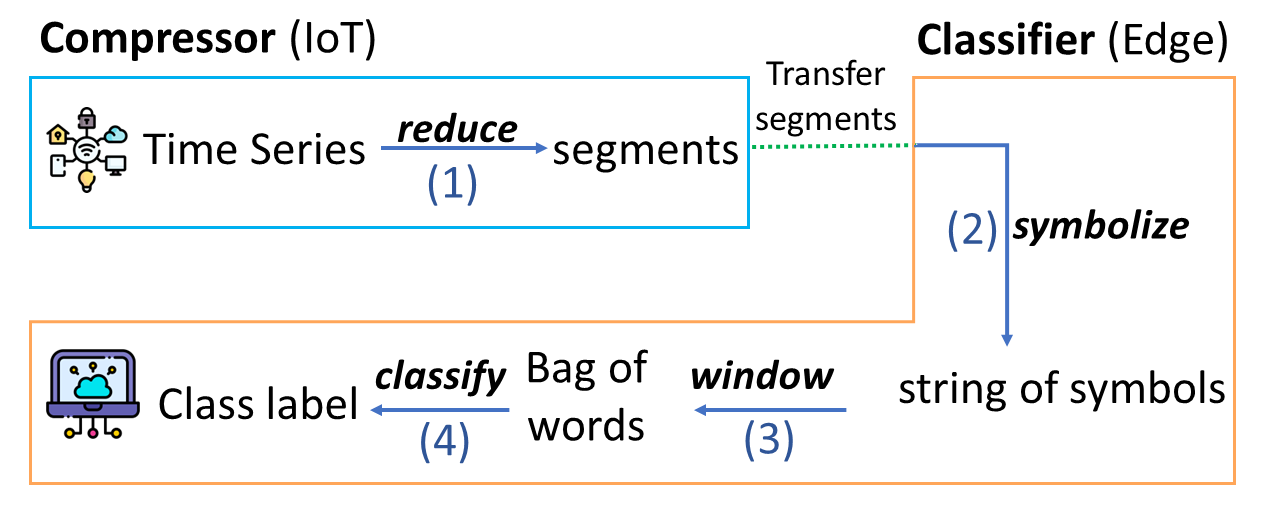}
    \vspace{-10pt} 
    \caption{A high-level view of our system model}
    \label{fig:overview}
    \vspace{-1\baselineskip}
\end{wrapfigure}
In this section, we provide a high-level overview of our approach, ABBA-VSM, as depicted in Figure~\ref{fig:overview}. ABBA-VSM has two main components, a \textbf{compressor} and a \textbf{classifier}.   The  \textbf{compressor} (e.g., IoT device) adaptively reduces TS up to linear segments (1) and transfers the resulting segments to Edge. The \textbf{classifier} (e.g., Edge device), receives the transmitted data,  encodes the linear segments as a string of symbols (2), and by applying the sliding window technique, it creates a bag of words (3), which we call as \textit{ABBA words},  and finally, it builds a classifier model to predict labels (4). 
\\
\textbf{Compressor:} Let us consider IoT devices as compressors. The IoT device sources a TS, denoted as 
\begin{equation}\label{eq:original_ts}
    T = [(x_1,y_1), ..., (x_N, y_N)] 
\end{equation}
where $x_i$ and $y_i$ are values and timestamps, respectively. Then, it reduces the TS by creating a polygonal chain of linear pieces, where each piece is bounded by the timestamp length and squared Euclidean distance error, according to the compression step in ABBA~\cite{elsworth2020abba}. These linear pieces are defined as $P = [p_1,p_2,...,p_n]$, where each linear piece $p_j =(len_j,inc_j)$ is a tuple of time step length and increment value, and with $n\leq N$. For simplicity, from now on we will refer to these linear pieces as segments. To mitigate misclassification issues, the compressor waits until all segments corresponding to a particular class are generated before sending a collection of segments to the classifier.  By creating such segments, the raw data  $T$ is converted into highly compressed data in the form of segments $P$. 
\\
\textbf{Classifier:} The classifier receives the segments $P = [p_1,p_2, ..., p_n]$ and clusters them to centers $C =[c_1,c_2,...,c_k]$ with $k\leq n$~\cite{elsworth2020abba}. Then, each cluster is symbolized using the alphabet $A = [a_1,a_2,...,a_k]$; thus, segments corresponding to the same cluster receive the same character as the cluster. 
Once we get the string of symbols representing the TS, we apply the windowing technique to create a \textit{bag of words}. Unlike traditional windowing methods that apply a sliding window to the original TS~\cite{schafer2015boss}, this paper proposes the application of a sliding window on compressed data, i.e., on a string of symbols. This approach facilitates the compression of raw TS at IoT without being influenced by sliding window characteristics (size and step). Then, we apply distance measure techniques to classify the labels of these bags of words. 

\section{Methodology} \label{sec:abba_vsm}
We have provided a high-level description of the proposed symbolic TSC in Section~\ref{sec:problem_formulation}. Here, we present detailed methodologies for constructing ABBA-VSM, consisting of the two main parts: \texttt{compressor} and \texttt{classifier}. For \texttt{compressor}, we use the Adaptive Brownian Bridge Aggregation (ABBA) technique, a continu-ous-time stochastic process that restricts Brownian motion to trajectories and converges to a given terminal state, enabling efficient data compression applications~\cite{elsworth2020abba}. By adaptively reducing the TS to linear segments, ABBA method demonstrates the ability to preserve the shape characteristics of TS better than other approaches like SAX and 1d-SAX representations. The key insight is that the segments can be modeled as a random walk with pinned start and end points, resembling a Brownian bridge. This allows for creating parameter-free (namely \textit{adaptive}) segments except for the choice of a reduction tolerance. Once the TS is reduced to segments, we transfer them to the Edge and encode them as a string of symbols. 

For \texttt{classifier}, we construct a Vector Space Model (VSM) using the string of symbols. VSM is a model that represents TS (in our case, symbolized TS) as vectors in a multi-dimensional space. A VSM  allows efficient similarity comparisons, from which a classification algorithm can be developed. Unlike classical Machine Learning models, which often require extensive training on large datasets (mainly on original TS) and involve significant computational power and memory, operations performed in VSM, such as similarity measurement, require fewer computational resources. A step-by-step description of methodology is described below and implemented in Algorithms~\ref{alg:training} and~\ref{alg:testing}.    
\begin{algorithm}[t!]
\caption{ABBA-VSM Training}
\label{alg:training}
\begin{algorithmic}[1]
\Procedure{ABBA-VSM Training}{training data, RT, C\_type, W\_size, W\_step}
    \State $corpus \gets$ empty\_list
    \While{True}
        \State $sample \gets$ get\_data\_from(training data) \label{alg:sample}
        \State $segments \gets$ reduction(sample, RT) \label{alg:segments}
        \State $string \gets$ cluster(segments, C\_type) \label{alg:string}
        \State $windowed\_string \gets$ window(string, W\_size, W\_step) \label{alg:window}
        \State $ABBA\_words \gets$ create\_bag\_of\_words(W\_string) \label{alg:abba_words}
        \State $labelled\_ABBA\_words \gets$ label\_words\_by\_class(ABBA\_words) \label{alg:labelled_abba_words}
        \State append $labelled\_ABBA\_words$ to corpus \label{alg:corpus}
    \EndWhile
    \State $training\_weight\_matrix \gets$ apply\_tf\_idf\_vectorizer(corpus) \label{alg:training_weights}
\EndProcedure
\end{algorithmic}
\end{algorithm}
\vspace{0\baselineskip}

\begin{algorithm}[t!]
\caption{ABBA-VSM Testing}
\label{alg:testing}
\begin{algorithmic}[1]
\Procedure{ABBA-VSM Testing}{unlabeled data, RT, C\_type, W\_size, W\_step}
    \State $segments \gets$ reduction(unlabeled data, RT) \label{alg:testing_segments}
    \State $string \gets$ cluster(segments, C\_type) \label{alg:testing_string}
    \State $windowed\_string \gets$ window(string, W\_size, W\_step) \label{alg:testing_window}
    \State $ABBA\_words \gets$ create\_bag\_of\_words(W\_string) \label{alg:testing_abba_words}
    \State $testing\_weights \gets$ apply\_tf\_idf\_vectorizer(words) \label{alg:testing_weights}
    \State $class\_label \gets$ cosine\_similarity(training\_weights, testing\_weights) \label{alg:similarity}
\EndProcedure
\end{algorithmic}
\end{algorithm}

\subsection{Compressor: Transforming Numerical Time Series Into Segments}
A time series compression method at IoT level involves a single step: \textit{reduce}. 
\textbf{Reduce:} We start by considering the time series $T$ in line~\ref{alg:segments} of Algorithm~\ref{alg:training}. 
Then, ABBA adaptively partitions $T$ into $n$ segments $P$ with $n<N$ in line~\ref{alg:segments} Algorithm~\ref{alg:training}. Each segment in $P$ consists of two values, the length of the time steps of each segment as $len_i := x_i - x_{i-1} \geq 1$ and increment in value as $inc_i := y_i - y_{i-1}$. The reduced time series is defined as follows,
\begin{equation}\label{eq:reduced_ts}
    \tilde{T}  = [(len_1,inc_1), ..., (len_n, inc_n)]\in \mathbb{R} ^{2\times n}
\end{equation}
The method is an adaptive compression as the Euclidean distance between $T$ and $\Tilde{T}$ is bounded by a user-defined reduction tolerance (\textit{RT}) value~\cite{elsworth2020abba}. Reducing raw TS into segments helps reduce the communication cost between IoT and Edge. It is important to note that we choose to create segments ($P$) on the IoT device and generate symbols on the Edge device. This distribution of computational tasks is necessary because symbols are needed at the Edge device, and creating symbols from segments ($P$) requires running a clustering algorithm for all new segments generated, which might be computationally infeasible for many IoT devices. Nevertheless, segments are already highly compressed TS, significantly reducing the communication costs between the IoT and Edge devices.

\subsection{Classifier: A Symbolic Approach}
The computational complexity of ABBA compression is $O(N)$ where $N$ is the number of data points in $T$. While clustering operations are relatively efficient for Edge environments, they can be computationally intensive for resource-constrained IoT devices. Thus, clustering followed by symbolization and then the construction of \textit{training} and \textit{testing} are performed at \texttt{classifier} part of ABBA-VSM on the Edge. 
\\
\textbf{Symbolize:} Similar tuples from $\Tilde{T}$ in Equation~\ref{eq:reduced_ts} form clusters, each encoded as a single character, allowing all segments that belong to the same cluster to be assigned one symbol in line~\ref{alg:string} of Algorithm~\ref{alg:training}. For this, tuple values $(len_i, inc_i)$ are separately normalized by their standard deviations $\sigma_{length}$ and $\sigma_{inc}$, respectively. Based on the empirical observations we conducted, the classification accuracy error between various clustering methods was negligible. Thus, for final empirical evaluation, we consider a sorting-based~\cite{CG22a} and the k-means algorithm. K-means clustering requires the number of clusters to be known beforehand, whereas the sorting-based method is adaptive, using the user-defined clustering tolerance \textit{CT}~\cite{CG22a}.

Finally, each tuple in the sequence $\Tilde{T}$ in Equation~\ref{eq:reduced_ts} is replaced by the symbol of the cluster it belongs to, resulting in the string of symbols $S = [s_1 s_2 ... s_n] \in A^n$.

\noindent\textbf{ABBA-VSM Training:} To create \textit{ABBA words} from a string of symbols in $S$, we apply the sliding window technique as shown in line~\ref{alg:window} of Algorithm~\ref{alg:training}. Compared to the traditional sliding window technique that is applied on time series before the compression, we propose to use the sliding window on symbolically compressed TS, i.e., on the string of symbols. Such a technique allows the adaptive compression of the original TS, forming Brownian bridges, without being affected by the sliding window dimensions. The sliding window dimensions can be pre-defined by the user. 
A \textit{term} corresponding to one window defines a single \textit{ABBA word}, a collection of ABBA words from one training sample forms a \textit{labeled bag of words} in line~\ref{alg:labelled_abba_words}. Then, a set of bags form \textit{corpus} in line~\ref{alg:corpus}.  Once all samples in training data are encoded as bags of \textit{ABBA words}, we group the labeled words to corresponding class labels and create a \textit{weight matrix}. This matrix defines the weights of all words in a corpus and is built as follows:

(\textbf{a}) \textit{Representation:} Let $d_i$, $D$ be a document that represents an individual class and a corpus, respectively.

(\textbf{b}) \textit{Term Frequency(TF):} For each word $t_j$ in document $d_i$, the $TF_{ij}$ is the number of times $t_j$ appears in $d_i$, i.e., 
    \begin{equation*}
        TF_{ij}=\frac{\text{number of times } t_j \text{ appears in } d_i}{\text{total number of terms in } d_i}
    \end{equation*}
    
(\textbf{c}) \textit{Inverse Document Frequency (IDF):} The $IDF_j$ quantifies a word's importance by calculating the logarithmic ratio of the total number of documents $|D|$ to the number of documents that include the term $t_j$.
    \begin{equation*}
        IDF_j = log(\frac{|D|}{\text{number of documents containing } t_j})
    \end{equation*}   
    
(\textbf{d}) \textit{TF-IDF}: By taking the product of $TF_{ij}$ and $IDF_j$, we calculate the importance of term $t_j$ in document $d_i$ as $W_{ij} = TF_{ij} * IDF_j$.
    
(\textbf{e}) \textit{Vector Representation:} Now each document $d_i$ can be represented as vector $v_i = (W_{i_1}, W_{i_2}, ... , W_{i_{|T|}})$. 
The rows are individual ABBA words in the \textit{weight matrix} (line~\ref{alg:training_weights} of Algorithm~\ref{alg:training}), and columns represent the class labels. 

\noindent\textbf{Testing:} To classify an unlabeled TS, ABBA-VSM transforms it into a frequency vector $w$ using the same steps that were used for training lines~\ref{alg:testing_segments}-\ref{alg:testing_weights} of Algorithm~\ref{alg:testing}. Then, it computes \textit{cosine similarity} values between frequency vector $w$ and $v_i$, with $i$ representing weight vectors for different class labels: 
\begin{equation*}
    similarity(w, v_i)=\frac{w\cdot v_i}{\|w\| \cdot \|v_i\|}
\end{equation*}
The unlabeled TS is assigned to the class label whose vector has the highest cosine similarity value (line \ref{alg:similarity} in Algorithm~\ref{alg:testing}).
\begin{figure*}[h!]
    \vspace{-1\baselineskip}
    \centering
    \includegraphics[width=1\textwidth]{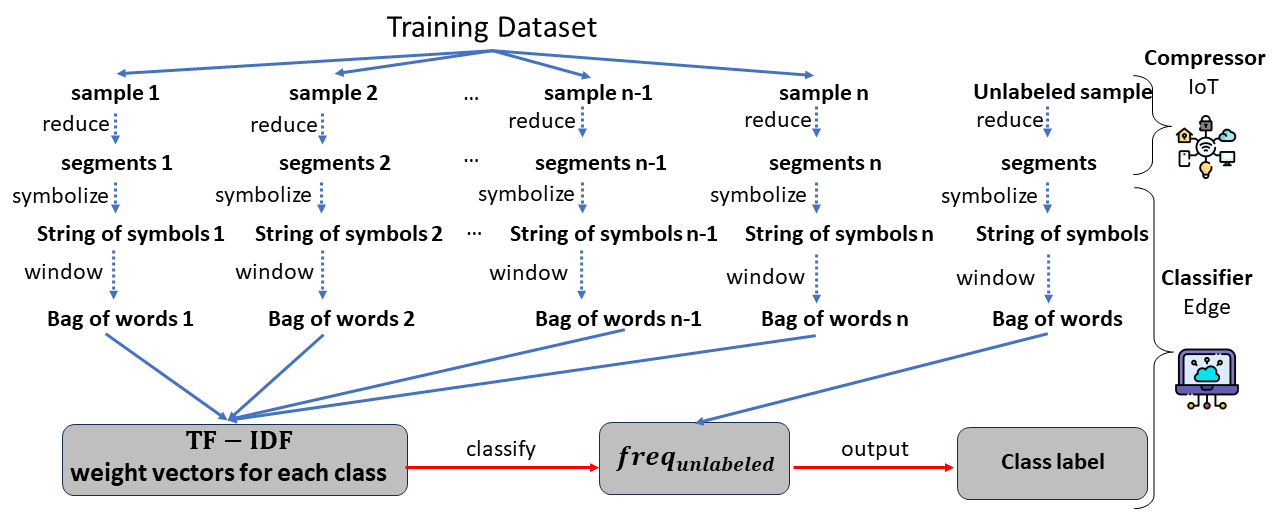}
    \caption{Building ABBA-VSM with training time series dataset and testing on unlabeled data.}
    \vspace{-1\baselineskip}
    \label{fig:methodology}
\end{figure*}

To further illustrate the whole process of symbolic classification, we provide a visual representation of the ABBA-VSM  in Figure \ref{fig:methodology}. As depicted, during the training phase, each TS sample is reduced and segmented by the ABBA method, with each segment being symbolized as a character to construct a string of symbols corresponding to the TS. Subsequently, this string of symbols is windowed to generate a bag of ABBA words. To build a classification model, the $TF-IDF$  approach is applied to each ABBA word, resulting in a weight matrix where rows represent individual ABBA words and columns denote class labels. For the classification of unlabeled samples, a similar technique is employed. A bag of ABBA words is built for a new sample, which is then transformed into a frequency vector. This vector is compared using cosine similarity to determine the class label. 

\section{Performance Evaluation} \label{sec:experiments}
In this section, we evaluate the performance of our proposed  ABBA-VSM,  a time series classiﬁcation algorithm based on approximating time series into symbols. We perform a range of experiments to assess its performance and to gain insights into both compression and classiﬁcation results.
\subsection{Metrics}
We present an evaluation framework for the ABBA-VSM algorithm. This framework encompasses a dual approach, incorporating both compression-based and classification-based metrics to assess the algorithm's performance across different dimensions. For compression-based metrics, we focus on quantifying the efficiency of the compression algorithm. First, the \textbf{compression ratio} (CR) measures the data reduction ratio achieved by the algorithm. A higher compression ratio implies more efficient compression, requiring less storage and providing potentially faster processing times. More formally, 
\begin{equation*}
    CR=1-\frac{Size \text{ } of \text{ } compressed \text{ } data}{Size \text{ } of \text{ } original \text{ } data}   
\end{equation*}
To overcome the complexity of storage measure in Python, we used a similar approach as in \cite{10.1007/978-3-031-39698-4_28} by assuming that the size of the original data is the length of the uncompressed float value multiplied by 4 (4 bytes to store numerical value), and the size of compressed data is the length of the compressed string (assuming 1 byte to store symbols/character). 
 
Conversely, for classification-based metrics, we focus on determining the algorithm's effectiveness in achieving \textbf{classification accuracy} (Acc) with compressed data.  A higher accuracy score signifies the algorithm's capability to maintain classification performance even with compressed data. Accuracy is calculated as:
\begin{equation*}
    Acc = \frac{Number \text{ } of \text{ } correctly \text{ } classified \text{ } samples}{Total \text{ } number \text{ } of \text{ } samples}\times 100\%
\end{equation*}

\subsection{Hyper-parameter Selection}
\begin{wraptable}{r}{6.5cm}
\vspace{0\baselineskip}
\caption{Hyperparameters and search space considered in the evaluation.}
\vspace{-1\baselineskip}
\label{tab:hyperparameter}
\scriptsize
\begin{tabular}{l l} \\ \hline
\textbf{Param.}  & \textbf{Search Space}      \\ \hline
\textit{RT} & \{0.001, 0.005, 0.01, 0.05, 0.1, 0.3, 0.5, 0.7\}  \\ \hline
\textit{Ctype} & \{k\_means, sorting\_based\} \\ \hline
\textit{CT} & \{0.001, 0.005, 0.01, 0.05, 0.1, 0.3, 0.5, 0.7\}  \\ \hline
\textit{Wsize} & \{2,3,4,5,6,7,8,9,10\} \\ \hline
\textit{Wstep}& \{1,2,3,4\}  \\  \hline
\textit{Csize} & \{2,3,4,5,6,7,8\} \\ \hline
\textit{Tsize} & \{0.05, 0.1, 0.2, 0.3, 0.4\} \\
\hline
\end{tabular}
\vspace{-1.8\baselineskip}
\end{wraptable}
We conducted an exhaustive experiment involving multiple hyperparameters, each playing a crucial role in shaping the performance of our classification model. These hyperparameters include \textit{reduction tolerance (RT)}, which establishes the threshold for considering data changes insignificant during compression; \textit{cluster type (Ctype)}, defining the clustering method for symbolic representation; \textit{clustering tolerance (CT)}, defining the dimension of cluster in sorting-based ABBA clustering; \textit{word size (Wsize)}, determining the length of ABBA words; \textit{window step (Wstep)}, specifying the stride for moving the sliding window along the compressed data;  \textit{cluster size (Csize)}, utilized in k-means based ABBA; and \textit{train-test split (Tsize)}, governing the proportion of samples allocated for model training and testing. Our study meticulously investigates the influence of these hyperparameter configurations on classification accuracy, with the ultimate goal of pinpointing the combination that optimally enhances accuracy. To ensure thorough exploration while avoiding excessive complexity, we define a well-structured search space for hyperparameters, encompassing a diverse range of values. This strategic selection of the search space strikes a balance between comprehensive exploration and computational feasibility, ensuring that our analysis yields robust and meaningful insights into the impact of hyperparameters on classification accuracy. Table \ref{tab:hyperparameter} lists the search space for each hyperparameter.

\subsection{Application Datasets}
\begin{wraptable}{r}{5 cm}
    \vspace{-2\baselineskip}
   \caption{Selected Time Series Datasets from UCR archive.}
   \vspace{-1\baselineskip}
    \label{tab:dataset}
    \scriptsize
    \begin{tabular}{lll} \\ \hline
    \textbf{Dataset}  & \textbf{\# classes} & \textbf{Shape} \\ \hline
    Adiac             & 37         & (731,1,176)      \\
    Beef              & 5          & (60,1,470)       \\
    CBF               & 3          & (930,1,128)      \\
    Coffee            & 2          & (56,1,286)      \\
    ECG200            & 2          & (200,1,96)      \\
    Face All          & 14         & (2250,1,131)     \\
    Face Four         & 4          & (112,1,350)    \\
    Fish              & 7          & (350,1,463)     \\
    Gun Point         & 2          & (200,1,150)    \\
    Lightning2        & 2          & (121,1,637)  \\
    Lightning7        & 7          & (143,1,319)   \\
    OliveOil          & 4          & (60,1,570)    \\
    OSU Leaf          & 6          & (442,1,427)   \\
    Syn. Control & 6          & (600,1,60)    \\
    Swedish  Leaf     & 15         & (1125,1,128)  \\
    Trace             & 4          & (200,1,275)   \\
    Two Patterns      & 4          & (5000, 1,128) \\
    Wafer             & 2          & (7164,1,152)  \\
    Yoga              & 2          & (3300,1,426) \\ \hline
    \end{tabular}
    \vspace{-6\baselineskip} 
\end{wraptable}

In our study, we analyze a variety of datasets from the UCR Time Series Classification Archive~\cite{UCRArchive2018}. These datasets are from different domains, such as classifying unicellular algae, examining beef spectrograms, and distinguishing between Robusta and Arabica coffee beans. Each dataset has varying numbers of classes, adding complexity to our analysis. For example, the Beef dataset includes five beef spectrograms, while the Fish dataset covers seven fish types. Table \ref{tab:dataset} provides an overview of 19 selected datasets used for experiments. We selected similar datasets as in~\cite{senin2013sax} to establish a comparison baseline. 

\subsection{Implementation}
To implement ABBA-VSM, we extended the fABBA 1.2.1 framework~\cite{CG22a} for the compression part, with modifications as required.Next, ABBA-VSM training and testing algorithms are implemented in Python and evaluated on an Intel(R) Core(TM) i7-8550U CPU @ 1.80GHz machine. We emulate the compressor (IoT device) and classifier (Edge device) locally on a single node.

\section{Results and Discussion} \label{sec:results}
In this section, we evaluate ABBA-VSM using the metrics defined in Section ~\ref{sec:experiments}. In addition, we compare our result with baselines and perform sensitivity analysis of hyperparameters. 

\subsection{Main Results}
\vspace{-2pt} 
\begin{wrapfigure}{l}{0.60\textwidth}
\vspace{-24pt}
 \includegraphics[width=0.59\textwidth]{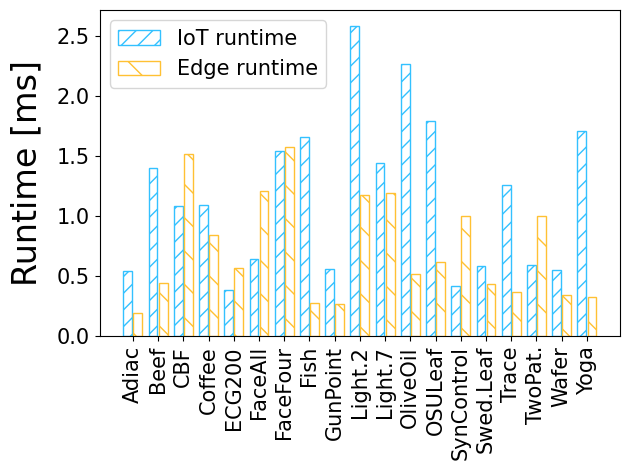}
    \vspace{-10pt}
        \caption{The total runtime overhead at compressor (IoT) and classifier (Edge) for each dataset.}
     \label{fig:latency_compression}
     \vspace{-7pt}
\end{wrapfigure}
\vspace{-2pt}

\begin{figure}[t!]
    \vspace{0\baselineskip}
    \centering
    \subfigure[Binary classification datasets]{
    \includegraphics[width=0.45\textwidth]{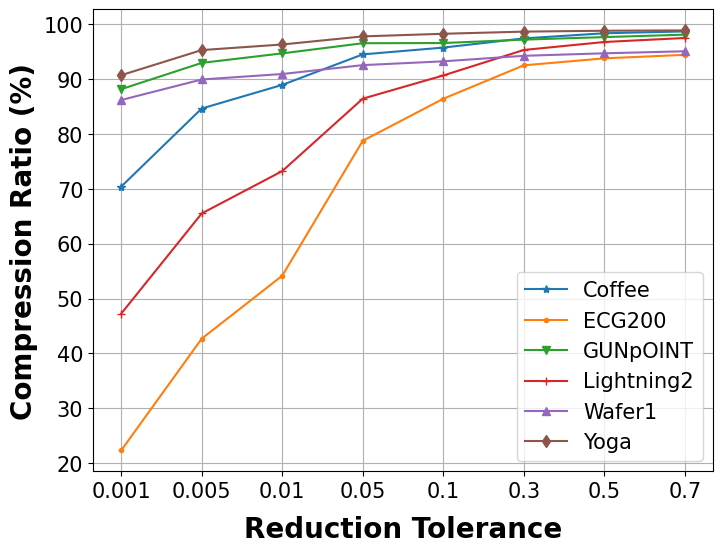}
    \label{fig:cr_binary}
    }
    \subfigure[Multiclass classification datasets]{
    \includegraphics[width=0.45\textwidth]{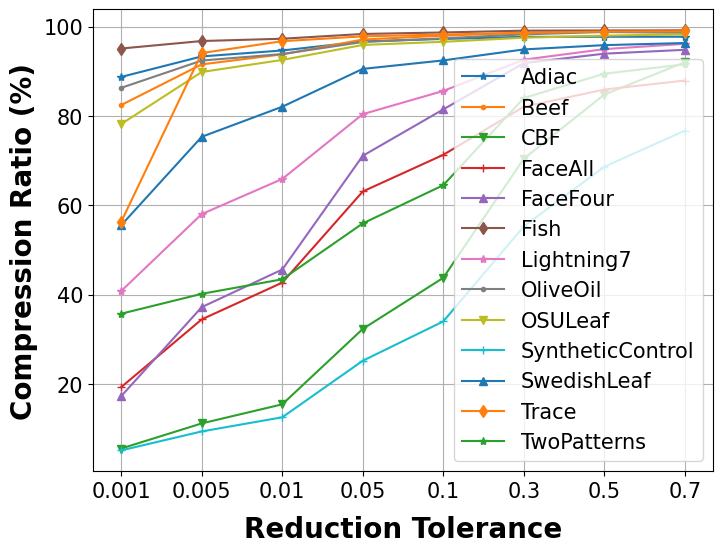}
     \label{fig:cr_multi}
  }
\caption{Average compression ratio with varying reduction tolerance. For binary classification datasets in (a) the compression ratio gradually increases with the increase of reduction tolerance. In contrast, multiclass data in (b) compression ratio converges to linear much faster.}
  \label{fig:compression_ratio}
  \vspace{-1\baselineskip}
\end{figure}

Our primary focus is to build the classification model using compressed data, emphasizing the need to achieve the best classification accuracy while maintaining a high compression ratio. Figure \ref{fig:compression_ratio} demonstrates the change in compression ratio with varying reduction tolerance. For both binary and multiclass classification datasets, we observe the increase in compression ratio with the increase in reduction tolerance. The average compression ratio for binary classification datasets is 80-90\%, whereas for multiclass, it is 50-60\%, shown in Figures  \ref{fig:cr_binary} and \ref{fig:cr_multi}, respectively.

As depicted in Figure~\ref{fig:latency_compression}, the runtime overhead of Edge is considerably lower compared to IoT. This disparity can be attributed to the Edge operating on already compressed time series (TS) data, while IoT processes the original uncompressed TS. 

Moreover, we compared the classification accuracy of our proposed ABBA-VSM model against several state-of-the-art baseline classifiers in~\cite{senin2013sax}, such as 1NN classiﬁers based on Euclidean distance and Dynamic Time Warping (DTW), Fast-Shapelets pattern, Bag of Patterns (BoP) and SAX-VSM. 1NN classifiers use Euclidean distance to provide a straightforward approach for classifying TS by measuring the closest neighbor in the feature space, while DTW additionally adapts to variations in TS length. Fast-Shapelets detect unique patterns in TS, focusing on the most significant segments. In contrast, the BoP converts TS into symbolic representations, which streamlines classification through the analysis of pattern frequencies. Additionally, SAX-VSM integrates SAX with VSM.

\textbf{The results demonstrate that our model outperforms all baseline methods on binary classification datasets}. For instance, on binary classification datasets such as ECG200, GunPoint, and Lightning2, ABBA-VSM achieved the highest performance with an accuracy of 100\%. In contrast, for Coffee and Wafer datasets, it performed as well as SAX-VSM by showing 90-100\% accuracy. In multi-class classification scenarios, the accuracy varies depending on the number of class labels; however, here, ABBA-VSM performed as well as baselines, with the exception of the Face All dataset, where it recorded the lowest accuracy at nearly 40\%. 

Table ~\ref{tab:error_rates} summarizes and compares the classification accuracy against the baselines given in~\cite{senin2013sax}.  In summary, our results demonstrate that ABBA-VSM is suitable in IoT and Edge environments for the development of Edge AI services utilizing classification services. 
\begin{table}[ht!]
\vspace{-1\baselineskip}
\caption{Comparison of classification accuracy (in the range of $[0,1]$, with 1 indicating 100\% accuracy) of ABBA-VSM against baselines from~\cite{senin2013sax}.}
\label{tab:error_rates}
\begin{center}
\scriptsize
\begin{tabular}{lllllllll}
\hline
\begin{tabular}[c]{@{}l@{}}\textbf{Dataset}\end{tabular}& 
\begin{tabular}[c]{@{}l@{}}\textbf{Type}  \end{tabular} & 
\begin{tabular}[c]{@{}l@{}}\textbf{1NN} \\ \textbf{Euclidian} \end{tabular} & 
\begin{tabular}[c]{@{}l@{}}\textbf{1NN} \\ \textbf{DTW} \end{tabular} & 
\begin{tabular}[c]{@{}l@{}}\textbf{Fast} \\ \textbf{Shapelets} \\ \textbf{Pattern} \end{tabular} &
\begin{tabular}[c]{@{}l@{}}\textbf{Bag} \\ \textbf{of} \\ \textbf{Patterns} \end{tabular} &
\begin{tabular}[c]{@{}l@{}}\textbf{SAX} \\ \textbf{VSM} \end{tabular} &
\begin{tabular}[c]{@{}l@{}}\textbf{ABBA-} \\ \textbf{VSM} \\ \textbf{sorting} \end{tabular} &
\begin{tabular}[c]{@{}l@{}}\textbf{ABBA-} \\ \textbf{VSM} \\ \textbf{k-means} \end{tabular} \\
\hline
Adiac            & multi & 0.61                   & 0.61              & 0.49     & 0.57                     & 0.62             & 0.5  & 0.45 \\
Beef             & multi & 0.53                   & 0.53              & 0.55     & 0.57                     & 0.97             & 1.00 & 1.00  \\
CBF              & multi & 0.85                   & 1.00              & 0.95     & 0.99                     & 1.00             & 0.79 & 0.7   \\
Coffee           &binary & 0.75                   & 0.82              & 0.93     & 0.96                     & 1.00             & 1.00 & 1.00   \\
ECG200           &binary & 0.88                   & 0.77              & 0.77     & 0.86                     & 0.86             & 1.00 & 1.00   \\
FaceAll          & multi & 0.71                   & 0.81              & 0.60     & 0.78                     & 0.79             & 0.39 & 0.27   \\
FaceFour         & multi & 0.78                   & 0.83              & 0.91     & 0.99                     & 1.00             & 1.00 & 1.00   \\
Fish             & multi & 0.78                   & 0.83              & 0.80     & 0.93                     & 0.98             & 1.00 &  1.00   \\
GunPoint        &binary & 0.91                   & 0.91              & 0.94     & 0.97                     & 0.99              & 1.00 & 1.00     \\
Lightning2       &binary & 0.75                   & 0.87              & 0.70     & 0.84                     & 0.80             & 1.00 & 0.90     \\
Lightning7       & multi & 0.57                   & 0.73              & 0.60     & 0.53                     & 0.70             & 0.9  &  0.79      \\
Olive Oil        & multi & 0.87                   & 0.87              & 0.79     & 0.77                     & 0.9              & 0.86  &  0.8     \\
OSU Leaf         & multi & 0.52                   & 0.59              & 0.64     & 0.76                     & 0.89             & 0.7  &  0.6    \\
Syn.Control      & multi & 0.88                   & 0.99              & 0.92     & 0.96                     & 0.99               & 0.67  &  0.6    \\
Swed.Leaf        & multi & 0.79                   & 0.79              & 0.73     & 0.80                     & 0.75             &  0.7     &  0.7    \\
Trace            & multi & 0.76                   & 1.00              & 1.00     & 1.00                     & 1.00             & 1.00  &  1.00      \\
Two Patterns      & multi & 0.91                   & 1.00              & 0.89     & 0.87                     & 1.00             &  0.9     &  0.86   \\
Wafer            &binary & 0.99                   & 0.98              & 1.00     & 1.00                     & 1.00             & 0.99  &  0.9     \\
Yoga             &binary & 0.83                   & 0.84              & 0.75     & 0.83                     & 0.84             & 0.71  &  0.67      \\
\hline
\end{tabular} 
\vspace{-2\baselineskip}
\end{center}
\vspace{-1\baselineskip}
\end{table}

\subsection{Sensitivity Analysis}
We present the outcomes of exhaustive experiments with various hyperparameters, each playing a crucial role in shaping the classification model's performance. To identify the relevant combination of hyperparameter values, we consider classification accuracy results above a threshold value of 80\%. 

\textbf{Reduction Tolerance:} An increase in reduction tolerance leads to a decrease in the number of segments, consequently increasing the compression ratio and reducing the storage demand. This raises the question of whether ABBA-VSM can achieve better classification accuracy with a higher compression ratio. Among all sets of hyperparameter value combinations for binary classification, the reduction tolerance value $RT=0.1$ achieved the accuracy threshold more frequently than the rest of the RT values. This is followed by tolerance ranges of $RT=\{0.3, 0.5\}$.  However, for multiclass datasets, the accuracy exceeding the threshold is achieved with even higher tolerances, such as $RT=\{0.3, 0.5, 0.7\}$, and can be seen in Figure \ref{fig:compression_ratio}. This evaluation shows us that RT can have a lower bound at 0.3 to achieve the accuracy threshold.

\textbf{Cluster Type: } In the binary classification case, the sorting-based clustering method outperformed k-means with an average accuracy improvement of 0.12, as seen in Table \ref{tab:error_rates}. However, both methods showed similar performance when dealing with multiclass datasets, with accuracy differences ranging only from 0.05 to 0.1. This suggests that the choice between sorting-based and k-means clusterings depends on the dataset's characteristics, especially on the number of classes involved.

\textbf{Test Size:} Nearly 80\% of datasets with more than 20\% test size performed lower than accuracy threshold. On the other hand,  almost 90\% of datasets with a test size less than 10\% showed accuracy between 80-100\%. This indicates that our method requires more data samples for training.

\textbf{Training and Test Time: } We observed that sorting-based clustering outperformed k-means clustering across all datasets in terms of computational efficiency. Specifically, when considering binary classification datasets, the k-means algorithm exhibited a notable increase in training time, ranging from 1.5x-2x more compared to sorting-based clustering. For multiclass datasets, k-means required approximately 0.5x-0.7x times more time than sorting-based clustering. For example, the FaceAll data required an average training time of 7.5 seconds for sorting-based clustering and 13.5 seconds for k-means-based clustering.  

\textbf{Window step and word size:} For both binary and multiclass datasets, we did not observe a significant classification accuracy change in different window steps and word sizes. Thus, to reduce the computational time, larger values for window step and word size, i.e., within the range $\{3,4\}$ and $\{7,8,9,10\}$, can be considered, respectively. 

\textbf{Clustering tolerance:} This hyperparameter is active when using \textit{sorting-based} clustering selection. While half of the binary classification datasets achieved the accuracy threshold with values between $CT=\{0.001, 0.005, 0.01, 0.05, 0.1\}$, the remaining half achieved with $CT=\{0.1, 0.3, 0.5, 0.7\}$. In approximately 80\% of multiclass datasets, the accuracy threshold is not met. This discrepancy can be attributed to the limited size of the training dataset, as indicated by experimental findings. 

\section{Conclusion and Future Works} \label{sec:conclusion}
We proposed ABBA-VSM, a symbolic time series classification method designed for resource-constrained Edge environments. Our proposed approach trains a Vector Space Model (VSM) classification model on a bag of words
(i.e., compressed and symbolized time series data)
and tests the model on unlabeled symbolic data. ABBA-VSM demonstrates 90-100\%  classification accuracy on binary classification datasets while achieving up to 80\% compression ratio.  In the future, we plan to explore the ABBA-VSM model for multivariate time series. 

\begin{credits}
\subsubsection{\ackname} This research has been partially funded through the projects: Transprecise Edge Computing (Triton), Austrian Science Fund (FWF), DOI: 10.55776/ P36870; Trustworthy and Sustainable Code Offloading (Themis), FWF, DOI: 10.55776/ PAT1668223 and Satellite-based Monitoring of Livestock in the Alpine Region (Virtual Shepherd) funded by Austrian Research Promotion Agency (FFG) Austrian Space Applications Programme (ASAP) 2022 \# 5307925.

\end{credits}

%
%
%
\bibliographystyle{splncs04}
%
\bibliography{mybibfile}

\end{document}